\documentclass[sigconf, nonacm=True]{acmart}

\usepackage{amsmath}
\usepackage{mathtools}
\usepackage{bbm}
\usepackage{color}

\newcommand{\note}[1]{}

\begin{document}

\title{Inherent Trade-offs in the Fair Allocation of Treatments}


\author{Yuzi He}
\additionalaffiliation{
\institution{University of Southern California}
\department{Department of Physics and Astronomy}
\city{Los Angeles}
\state{CA}
\country{USA}
}
\affiliation{
\institution{USC Information Sciences Institute}
\city{Marina del Rey}
\state{CA}
\country{USA}
}
\email{yuzihe@usc.edu}

\author{Keith Burghardt}
\affiliation{
\institution{USC Information Sciences Institute}
\city{Marina del Rey}
\state{CA}
\country{USA}
}
\email{keithab@isi.edu}

\author{Siyi Guo}
\additionalaffiliation{
\institution{University of Southern California}
\department{Department of Computer Science}
\city{Los Angeles}
\state{CA}
\country{USA}
}
\affiliation{
\institution{USC Information Sciences Institute}
\city{Marina del Rey}
\state{CA}
\country{USA}
}
\email{siyiguo@usc.edu}

\author{Kristina Lerman}
\affiliation{
\institution{USC Information Sciences Institute}
\city{Marina del Rey}
\state{CA}
\country{USA}
}
\email{lerman@isi.edu}

\begin{abstract}
Explicit and implicit bias clouds human judgement, leading to  discriminatory treatment of minority groups. A fundamental goal of algorithmic fairness is to avoid the pitfalls in human judgement by learning policies that improve the overall outcomes while providing fair treatment to protected classes. In this paper, we propose a causal framework that learns 
optimal intervention policies from data subject to fairness constraints. We define two measures of treatment bias and infer best treatment assignment that minimizes the bias while optimizing overall outcome.  We demonstrate that there is a dilemma of balancing fairness and overall benefit; however, allowing preferential treatment to protected classes in certain circumstances (affirmative action) can dramatically improve the overall benefit while also preserving fairness. 
We apply our framework to data containing student outcomes on standardized tests and 
show how it can be used to design real-world 
policies that fairly improve student test scores.  
Our framework provides a principled way to learn fair treatment policies in real-world settings.
\end{abstract}

\maketitle

\section{Introduction}

Equitable assignment of treatments is a fundamental problem of fairness, especially in cases where treatments are not available to all people and some individuals stand to benefit more from them than others. This problem arises in multiple contexts, including allocating costly medical care to sick patients~\cite{rajkomar2018ensuring,emanuel2020fair}, materials following a disaster~\cite{wang2019measuring}, college spots and financial aid in college admissions, extending credit to consumers, and many others. Despite growing interest from the research community~\cite{elzayn2019fair,nabi2019learning,donahue2020fairness} and the rise of  automated decision support systems in healthcare and college admissions that help make such decisions~\cite{obermeyer2019dissecting}, fair allocation of scarce resources and treatments remains an important open problem.

To motivate the problem, consider an infectious disease, like the COVID-19 pandemic, spreading through the population. The toll of the pandemic varies in different ethnic and racial groups (i.e., \textit{protected} groups) and also via several comorbidities, such as age, weight, underlying medical conditions, etc. 
When a vaccine becomes available, who should receive it first? 
To minimize loss of life, we could reserve the vaccine 
for high-risk groups with comorbidities, but this does not 
guarantee protected groups will be treated equitably. Some groups will get preferential treatment, unless---and this is highly unlikely---all groups reside in high-risk categories at equal rates. 
In comparison, adding 
fairness constraints to provide more vaccines to protected groups 
may result in more lives lost overall, in cases where protected groups have lower mortality. 
This demonstrates the difficult trade-offs policy-makers must consider regardless of the policies they choose.

Similar trade-offs between unbiased and optimal outcomes often appear in automated decisions. 
This issue received much attention since an investigation by ProPublica found that software used by judges in sentencing decisions was systematically biased~\cite{angwin2016machine}. The software’s algorithm deemed Black defendants to be a higher risk for committing a crime in the future than White defendants with similar profiles. Subsequent studies showed the cost in making the algorithm less biased is a decrease in its accuracy~\cite{corbett2017algorithmic,menon2018}. 
As a result, a fairer algorithm is more likely to incorrectly label violent offenders as low-risk and vice versa. This can jeopardize public safety if high-risk offenders are released, or needlessly keep low-risk individuals in jail. 

We will also show that there are multiple ways to define fair policies which do not necessarily overlap. Going back to the vaccine example, selecting individuals from the population at random to receive the limited doses of the vaccine may be considered equitable, but there might be grave differences in mortality rates across protected classes, which this treatment would not overcome. In contrast, preferentially giving vaccines to protected groups may create equitable losses of life between classes, but implies unfair allocation of resources  and will not benefit the population the most. Kleinberg et al. made a similar finding for automated decisions~\cite{kleinberg2016inherent}. 
 Except for rare trivial cases, a fair algorithm cannot simultaneously be balanced (conditioned on outcome, predictions are similar across groups)~\cite{angwin2016machine} and well-calibrated (conditioned on predictions, the outcomes will be similar across groups)~\cite{kleinberg2016inherent}. Decreasing one type of bias necessarily increases the other type. Empirical analysis confirmed these trends in benchmark data sets~\cite{he2020geometric}. More worrisome still, there are dozens of definitions of AI fairness~\cite{Verma2018}, and we suspect there is also no shortage of fair policy definitions, making an unambiguous definition of ``fair'' a challenge. 



In the current paper, we combine causal inference with fairness to learn optimal treatment policies from data that increase the overall outcome for the population. First, we define novel metrics for fairness in causal models that account for the heterogeneous effect a treatment may have on different subgroups within population. These metrics measure \textit{inequality of treatment opportunity} (who is selected for treatment) and \textit{inequality of treatment outcomes} (who benefits from treatment). 
This compliments previous research to maximize 
utilization of resources, i.e., ensuring that they do not sit idle, while also maximizing fairness 
\cite{elzayn2019fair}. 

We also show a necessary trade-off between fair policies and those that provide the largest benefit of treatment to the most people. We then show how affirmative action policies that preferentially select individuals from protected subgroups for treatment can improve the overall benefit of the treatment to the population, for a given level of fairness. Thus we find a necessary trade-off between policies that are fair overall with policies that would be fair within subgroups. These results demonstrate novel ways to improve fairness of treatments, as well as the important trade-offs due to distinct definitions of fairness. 

Our methods are tested in synthetic and real-world data. Using high school student test scores and school funding in different regions of the US, we devise fair funding policies that reduce discrimination against counties with a high percentage of Black families. 
Because the protected subgroup is more sensitive to the treatment (school funding), we create an affirmative action policy in which school funding tends to increase in regions with more Black families. This policy could raise test score more fairly than alternative funding policies.

The rest of the paper is organized as follows. We begin by reviewing related work, then we describe the causal inference framework we use to estimate the heterogeneous effect of treatments, define treatment biases, and optimization algorithm that learns fair intervention policies from data. We explore the methods in synthetic data, as well as real-world data.

\section{Related Work}
Trade-off between fairness and optimal prediction is intuitively unavoidable. We can regard fairness condition as a constraint to the optimization and the optimal solution which satisfies the constraint will be a sub-optimal. In our case, this means that when designing the intervention policy, we have to sacrifice overall benefit in order to make our policy fair. 

Fairness is first considered in predictions and representations. Early works include constrained Logistic regressions proposed by Zafar et al. \cite{zafar2017fairness, zafar2017fairnessbeyond, zafar2017parity}.
Menon et al. \cite{menon2018} related two fairness measures, disparate impact (DI) factor and mean difference (MD) score to cost sensitive fairness aware learning. 
There has also been extensive research on autoencoder based method which produced fair representation (embedding) \cite{moyer2018invariant, louizos2015variational} and generative models which generates data instances which are fair \cite{xu2019fairgan+,grover2019fair,kairouz2019censored}. 

Recently, there is a growing literature of fairness in causal inference, decision making and resource allocation. There are case studies on social work and health care policy such as \cite{chouldechova2018case, rajkomar2018ensuring, emanuel2020fair}. Corbett-Davies et al. \cite{corbett2017algorithmic} formulate the fair decision making as optimization problem under the constraints of fairness. This can be regarded as an easy adaption from fair prediction task such as \cite{zafar2017fairness}. Kusner et al. \cite{kusner2017counterfactual} proposed a new perspective of fairness based on causal inference, \textit{counterfactual fairness}. The counterfactual fairness requires the outcome be independent of the sensitive feature, or in other words, conditional on confounders, which further differs from 
\textit{equal opportunity} \cite{Hardt2016}, or still other metrics, such as the \emph{80\% rule}, \textit{statistical parity}, \emph{equalized odds}, or \emph{differential fairness} \cite{Foulds2019}.

Donahue and Kleinberg \cite{donahue2020fairness} studied the problem of fair resource allocation.
The goal was maximizing utility under the constrain of fairness, 
from which theoretical bound for the gap between fairness and unconstrained optimal utility is derived. 
Elzayn et al. \cite{elzayn2019fair} considered a similar case, with potentially more realistic assumptions that the actual demand is unknown and should be inferred.
The problem is formulated as constrained optimization, 
with the help of \textit{censored} feedback.

Zhang et al. \cite{zhang2016causal} modeled direct and indirect discrimination using path specific effect (PSE) and proposed a constrained optimization algorithm to eliminate both direct and indirect discrimination. 
Also based on the concept of PSE, Nabi et al. \cite{nabi2018fair} considered performing fair inference of outcome from the joint distribution of outcomes and features. 
Chiappa \cite{chiappa2019path} also proposed PSE based fair decision making by simply correcting the decision at test time.

Our work, however, differs from this previous work because we create (a) \emph{policy}-based definitions of fairness, (b) optimize on whow to treat while accounting for fairness trade-offs, and (c) address an under-explored trade-off between equal opportunity and affirmative action to improve policy fairness.

\section{Methodology}
We briefly review heterogeneous treatment effect estimation, which we use to learn fair treatment policies. We then discuss how we measure biases in treatments, and create optimal intervention strategies.

\subsection{Heterogeneous Treatment Effect }
\label{sec:hte}

Suppose we are given $N$ observations indexed with $i=1,...,N$, consisting of tuples of data of the form of $(X_i, y^{\text{obs}}_i, t_i)$. Here $X$ denotes features of the observation, $y^{\text{obs}}$ is the observed outcome, and binary variable $t$ indicates whether the observation came from the \textit{treated group} ($t = 1$) or the \textit{control} ($t = 0$). We assume that each observation $i$ has two potential outcomes: the controlled outcome $y_i^{(0)}$  and the treated outcome $y_i^{(1)}$, but we only observe one outcome $y^{\text{obs}}_i = y_i^{(t_i)}$. 
In addition, we assume that given features $X$, both of the  potential outcomes $y^{(0)}, y^{(1)}$ are independent of the treatment assignment $t$.
\begin{equation*}
    (y^{(0)}, y^{(1)})\perp t | X.
    \label{eq:unconf}
\end{equation*}
This condition is called the \textit{unconfoundedness assumption}.

The heterogeneous treatment effect is defined as 
\begin{equation}
    \tau(X) = \mathbb{E}[y^{(1)} - y^{(0)} | X]
    \label{eq:treatment}
\end{equation}
The task of heterogeneous treatment effect (HTE) estimation is to construct an optimal estimator $\hat \tau(X)$ from the observations. A standard model of HTE is a causal tree~\cite{athey-pnas16}. Causal trees are similar to classification and regression trees (CART), as they both rely on recursive splitting of the feature space $\mathbf{X}$, but causal trees are designed to give the best estimate of the \textit{treatment effect}, rather than the outcome. To avoid overfitting, we employ an honest splitting scheme~\cite{athey-pnas16}, 
in which half of the data is reserved to estimate the treatment effect on leaf nodes. The objective function to be maximized for honest splitting is the negative expected mean squared error of the treatment effect $\tau$, defined as below.
\begin{multline}
    -\widehat{\text{EMSE}}_{\tau}(S^{\text{tr}}, N^{\text{est}}, \Pi) = \frac{1}{N^{\text{tr}}}\sum_{i\in S^{\text{tr}}}\hat{\tau}^2(X_i;S^{\text{tr}}, \Pi)\\
    - (\frac{1}{N^{\text{tr}}}+\frac{1}{N^{\text{est}}})\cdot \sum_{l\in \Pi}(\frac{\mathrm{Var}^{\text{tr}}(l|t=1)}{p} + \frac{\mathrm{Var}^{\text{tr}}(l|t=0)}{1-p})
    \label{eq:honest_tree}
\end{multline}
Here $S^{\text{tr}}$ is the training set, $N^{\text{tr}}$ and $N^{\text{est}}$ are the size of training and estimation set. $\Pi$ is a given splitting, $l$ is a given leaf node, $p$ is the ratio of data being treated. Terms $\mathrm{Var}^{\text{tr}}(l|t=0)$ and $\mathrm{Var}^{\text{tr}}(l|t=1)$ are the within-leaf variance calculated for controlled and treated data on training set. Note that we only use the size of the estimation data during splitting. In cross validation, we use the same objective function and plug in validation set $S^{\text{val}}$ instead of $S^{\text{tr}}$.

After a causal tree is learned from data, observations in each leaf node correspond to groups of similar individuals who experience the same effect, in the same way a CART produces leaf nodes grouping similar individuals with similar predicted outcomes. 

\subsection{Inequalities in Treatment}
\label{sec:method_bias}
In many situations of interest, data comes from a heterogeneous population that includes some protected subgroups, for example, racial, gender, age, or income groups. We categorize these subgroups into one of $k$ bins, $z \in [1, k]$
.  Even though we do not use $z$ as a feature in HTE estimation, the biases present in data may distort learning, and lead to to infer policies that unfairly discriminate against protected subgroups. 

An additional challenge in causal inference is that a treatment can affect the subgroups differently. 
To give an intuitive example, consider a hypothetical scenario where a high school is performing a supplemental instruction program (intervention or treatment) to help 
students who are struggling academically.
Student are described by features $X$, such as age, sex, race, historical performance, average time spent on homework and computer games, etc. We want our intervention to be fair with respect to students with different races (in this case, the sensitive feature $z$ is race). That means we may want to both reduce the performance gap between different races and we also want to make sure that the minority race gets ample opportunity to participate in the intervention program. However, we assume that the school district has limited resources for supplemental instruction, which means that not every struggling student can be assigned to the intervention program. 
To best improve the overall performance, it therefore makes sense to leave more spots in the program to 
students who are more sensitive to intervention (they have a large treatment effect $\tau(X)$). But if the previous pilot programs show that the effect of the intervention is different amount subgroups (e.g., races), with one subgroup more sensitive to the intervention and also having a better average outcome, we have a dilemma between optimal performance and fairness. \note{dilemma explained.} If we only care about optimal outcome, the intervention will lead to not only larger performance gap between races but also lack of treatment opportunity for minority race. If we assign the intervention randomly, we will not make full use of the limited resource to benefit the population. 



Below we discuss our approach to measure bias in treatment or intervention. We learn the effect of the interventions using causal trees.
A causal tree learned on some data partitions individual observations among the leaf nodes. A group of $n_i$ observations associated with a leaf node $i$ of the causal tree is composed of $n_i^{(1)}$ observations of treated individuals and $n_i^{(0)}$ controls. We can further disaggregate observations in group $i$ based on the values of the sensitive attribute $z$. This gives us $n_{i,z=j}$ as the size of subgroup $z=j$, which has $n^{(1)}_{i,z=j}$ treated individuals and $n^{(0)}_{i,z=j}$ controls. Similarly if $y^{(0)}_i$ and $y^{(1)}_i$ are the estimated outcomes for the control and treated individuals in group $i$, then $y^{(0)}_{i,z=j}$ and $y^{(1)}_{i,z=j}$ as the estimated outcomes for the control and treated subgroup $z=j$ in group $i$. Table~\ref{tab:definitions} lists these definitions.

\begin{table}[ht]
    \centering
    \begin{tabular}{p{5.5cm}|l}
    \hline
        number of individuals in leaf node $i$ &  $n_i$\\
        number of control/treated individuals in group $i$ &  $n_i^{(0)}$, $n_i^{(1)}$ \\
        number of individuals in subgroup $z=j$ &  $n_{i,z=j}$ \\
        number of control/treated individuals in subgroup $z=j$ & $n^{(0)}_{i,z=j}$, $n^{(1)}_{i,z=j}$ \\
        outcome in the control/treated group & $y^{(0)}_i$, $y^{(1)}_i$ \\
        outcome for control/treated subgroup $z=j$ & $y^{(0)}_{i,z=j}$, $y^{(1)}_{i,z=j}$ \\ 
        \hline
        treatment ratio for leaf node $i$ & $r_i$\\
        treatment ratio for subgroup $z=j$ in leaf node $i$ & $r_{i,z=j}$ \\
        overall outcome & $\bar y$\\
         \hline
        maximum number of treated individuals   & $N_{\max}^{(1)}$\\
        maximum ratio of treated individuals   & $r_{\max}$\\
        inequality of treatment opportunity & $\text{Bias}_r$ \\
        inequality of treatment outcomes & $\text{Bias}_y$ \\ \hline
    \end{tabular}
    \caption{Definitions used in measuring biases in treatment.}
    \label{tab:definitions}
\end{table}

\subsubsection{Measuring Inequality of Treatment Opportunity}
To quantify the inequalities of treatment, we first look at  the \textit{inequality of treatment opportunity}, i.e., 
the disparity 
of the assignment of individuals from the protected subgroup in leaf node $i$ to the treatment condition.
To measure the bias, we introduce the treatment ratio $r_{i,z=j}$ as the fraction of treated individuals from subgroup $j$ among the group in leaf node  $i$:
$$r_{i,z=j} = \frac{ n_{i,z=j}^{(1)}}{n_{i,z=j}^{(0)} + n_{i,z=j}^{(1)}}.$$ We define the inequality of treatment opportunity as the maximum difference of the within-leaf treatment ratios taken over all leaf nodes $i$ and pairs of 
subgroups $j,\ j'$,
\begin{equation}
    \text{Bias}_r = \max_{i,j,j'} |r_{i,z=j} - r_{i,z=j'}|.
\end{equation}

\begin{figure}
    \centering
    \includegraphics[width=0.9\linewidth]{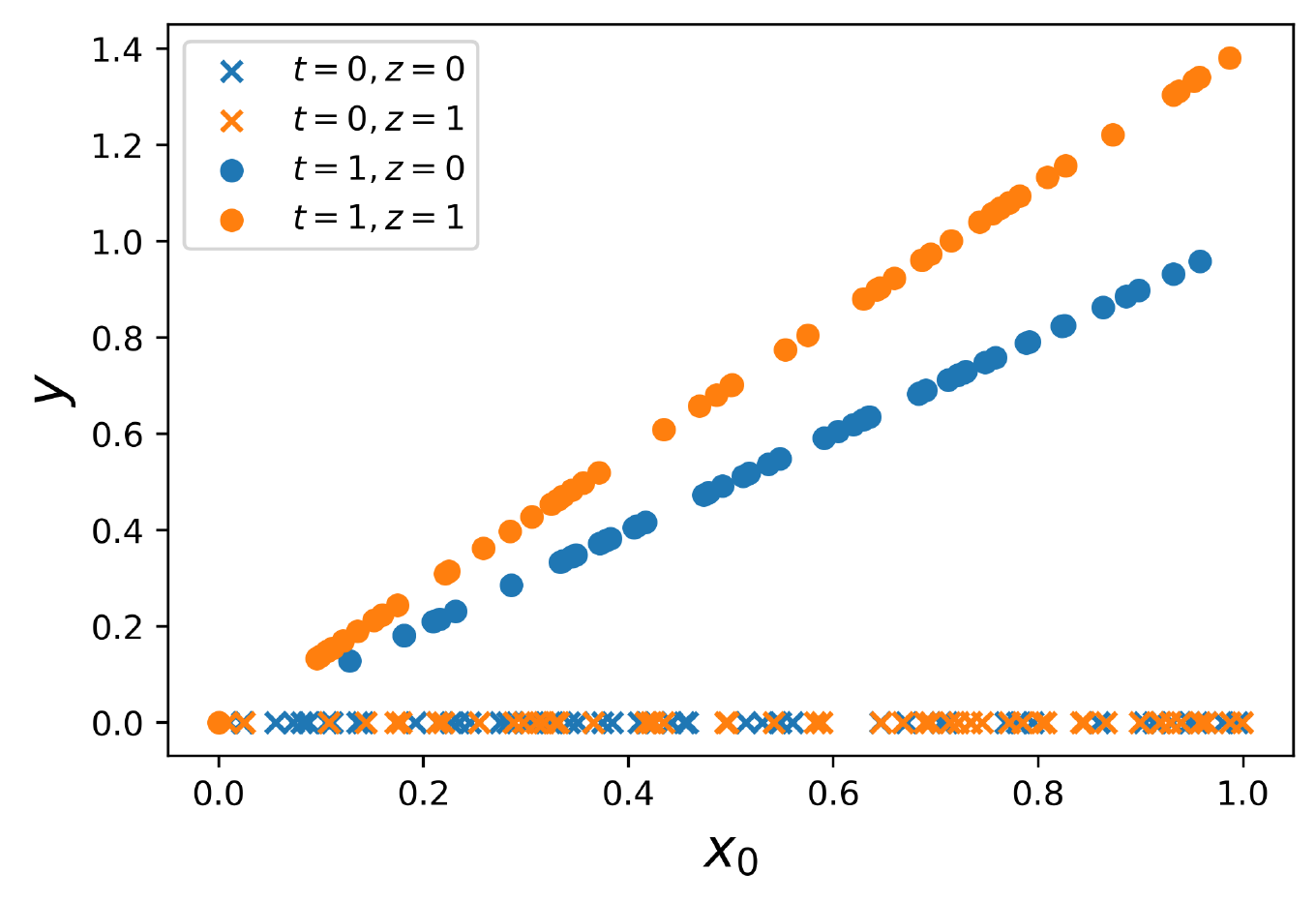}
    \caption{The outcome $y$ vs feature $x_0$ plot for synthetic data. Note that the other feature $x_1$ is independent from $y$.}
    \label{fig:syn_data}
\end{figure}

\begin{figure}[tbh!]
    \centering
    \begin{minipage}{0.90\linewidth}
        \includegraphics[width=\linewidth]{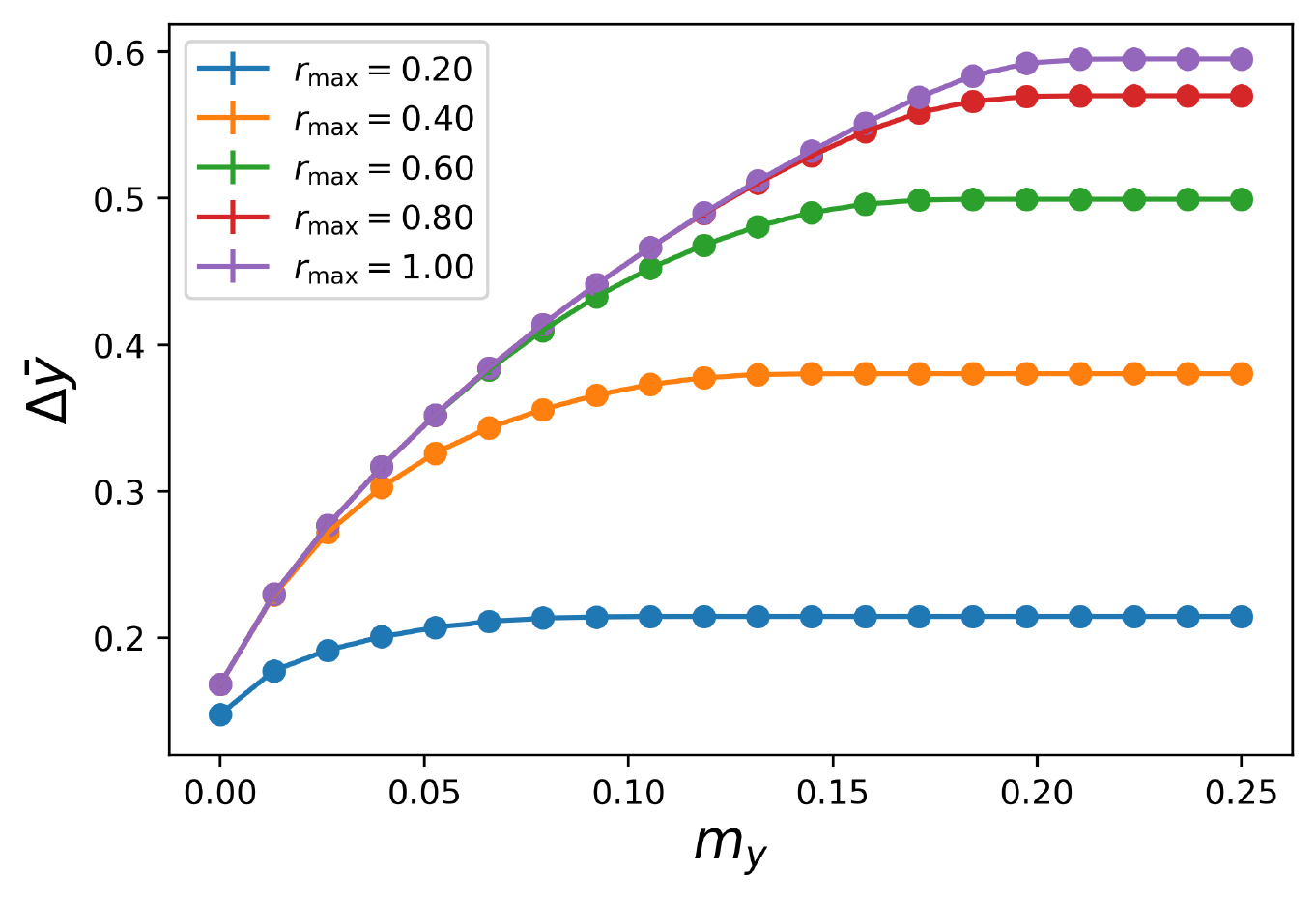}
        \caption*{(a) Equal opportunity}
    \end{minipage}
    \begin{minipage}{0.90\linewidth}
        \includegraphics[width=\linewidth]{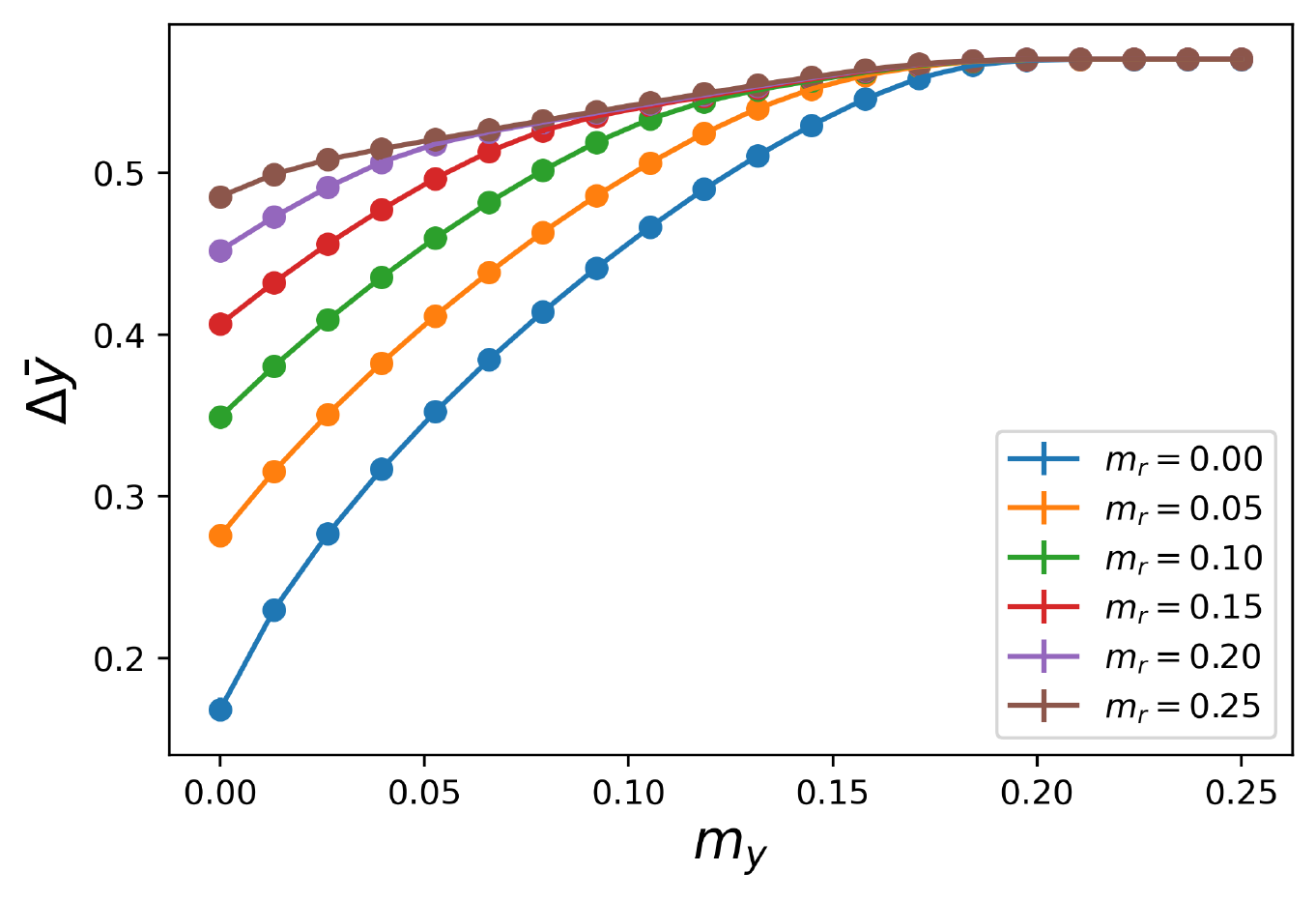}
        \caption*{(b) Affirmative action}
    \end{minipage}    
    \caption{
    Overall improvement in outcome after treatment as a function of the discrepancy of outcomes in the protected groups. (a) $\Delta \bar y$ vs $m_y$ when equal treatment opportunity is assumed. Different curves show the efficient boundary of  policies under constrains of certain limit of resource $r_{\max}$ and maximum allowed bias $m_y$. (b) $\Delta \bar y$ vs $m_y$ when affirmative action is allowed. Here $r_{\max} = 0.8$ and different curves shows different degrees of affirmative action, measured by $m_r$. Affirmative action greatly improves $\Delta \bar y$ in case where $m_y$ is low, or constrain is ``tight.''}
    \label{fig:syn_bound}
\end{figure}

\subsubsection{Measuring Inequality of Treatment Outcomes}
The second type of bias we measure is the \textit{inequality of treatment outcomes}. 
This bias arises because subgroups may differ in their 
response to treatment and their controlled outcomes. We quantify this disparity as 

\begin{equation}
    \bar y_{z=j} = \frac{1}{\sum_i n_{i,z=j}}\sum_i n_{i,z=j}\cdot \left [ r_{i,z=j}\cdot y_{i,z=j}^{(1)} + (1-r_{i,z=j})\cdot y_{i,z=j}^{(0)} \right ],
\end{equation}
where the index $i$ is for leaf nodes of the causal tree. We define inequality of outcomes as the largest difference of expected outcomes for all pairs of protected subgroups 
\begin{equation}
    \text{Bias}_y = \max_{j,j'}|\bar y_{z=j} - \bar y_{z=j'}|.
\end{equation}
\noindent Note that when there are only two protected groups, it is not necessary to take the maximum.

\subsection{Learning Optimal Interventions}
\label{sec:method_opt}
A crucial problem in the design of interventions is how to balance between the optimal performance and bias.
Below we describe learning optimal interventions 
that maximize the overall benefit of treatment while properly control the bias of treatment opportunity and the bias of outcome among different subgroups. We can achieve optimality  by choosing which individuals to treat. 
Specifically, given the features $X$, the potential outcomes $y^{(0)}$ and $y^{(1)}$ are independent of treatment assignment $t$. Therefore, we can vary $r_{i,z=j}$, while keeping $y_{i,z=j}^{(0)}$ and $y_{i,z=j}^{(1)}$ constant, as part of the optimal policy.

\subsubsection{Equal Treatment Opportunity-Constrained Interventions}
As a first step, let us consider the case in which $\text{Bias}_r = 0$, i.e., all subgroups have the same fraction of treated individuals, 
and the equality of treatment opportunity is strictly satisfied. For every group $i$, we assign the same treatment ratio $r_i$ to all subgroups within $i$ defined by $z$. The mean of overall outcome can be written as 
\begin{equation} 
    \bar y = \frac{1}{\sum_i n_i} \sum_i n_{i} \cdot \left [ 
    r_i\cdot y_{i}^{(1)} + (1-r_i)\cdot y_{i}^{(0)}
    \right].
\end{equation}
Our objective is to maximize $\bar y$ by varying $r_i$, subject to the following constraints:
\begin{itemize}
    \item First we set an upper bound for the inequality of outcomes, meaning we will not tolerate a disparity in outcomes that is larger than $m_y$,
    \begin{equation}
        \text{Bias}_y \leq m_y
    \end{equation}
    
    \item Practically speaking, the treatment is often bounded by the availability of resources, which usually means that we can treat at most  $N_{\text{max}}^{(1)}$ people,
    \begin{equation}
        \sum_i n_i\cdot r_i \leq N_{\text{max}}^{(1)}
    \end{equation}
    
    \item Finally, treatment ratios have to satisfy a trivial  constraint, 
    \begin{equation}
        0 \leq r_i \leq 1
    \end{equation}
\end{itemize}

\begin{figure*}[tbh]
    \centering
    \begin{minipage}{0.24\linewidth}
        \includegraphics[width=\linewidth]{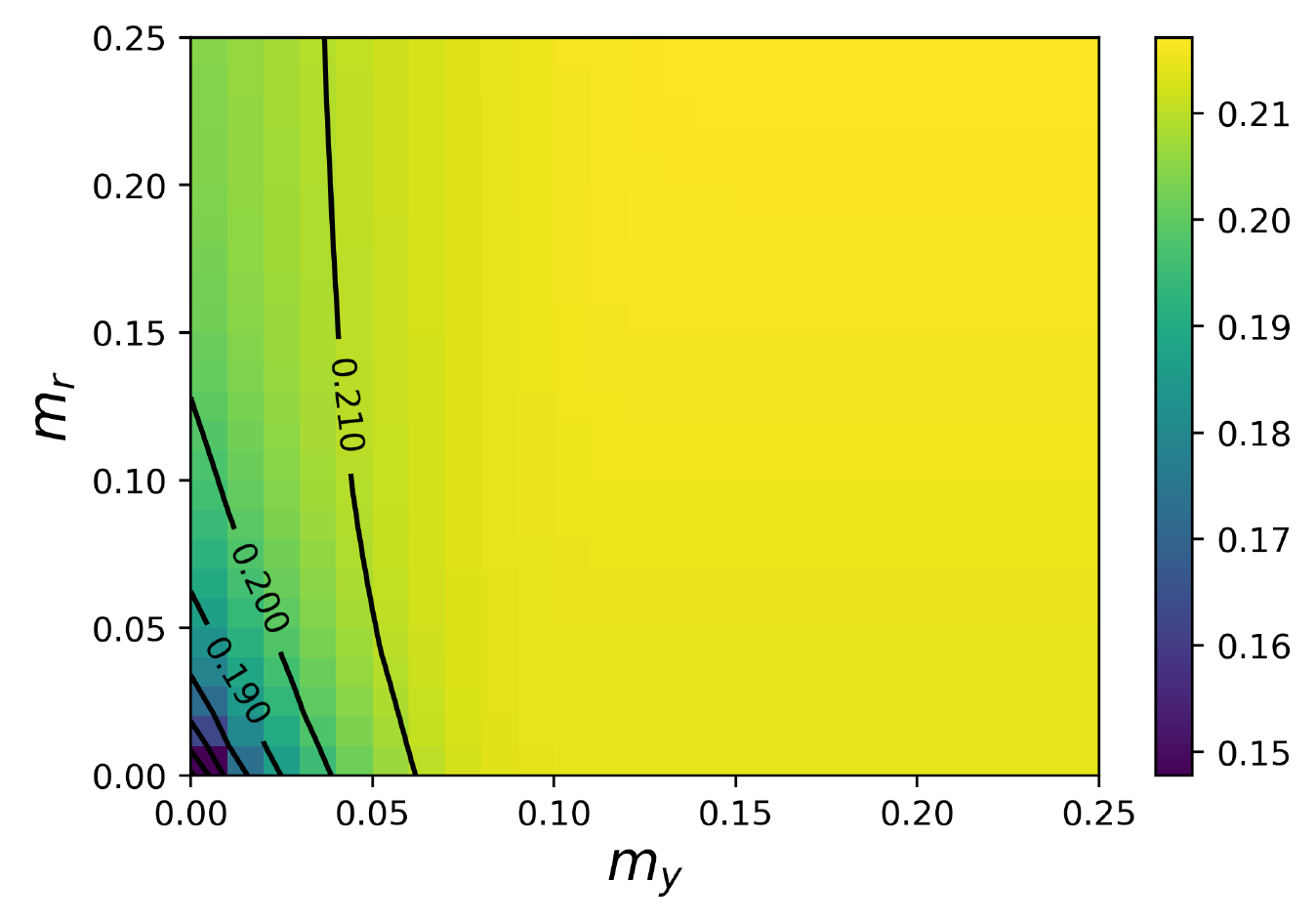}
        \caption*{(a) $r_{\max}=0.2$ }
    \end{minipage}
    \begin{minipage}{0.24\linewidth}
        \includegraphics[width=\linewidth]{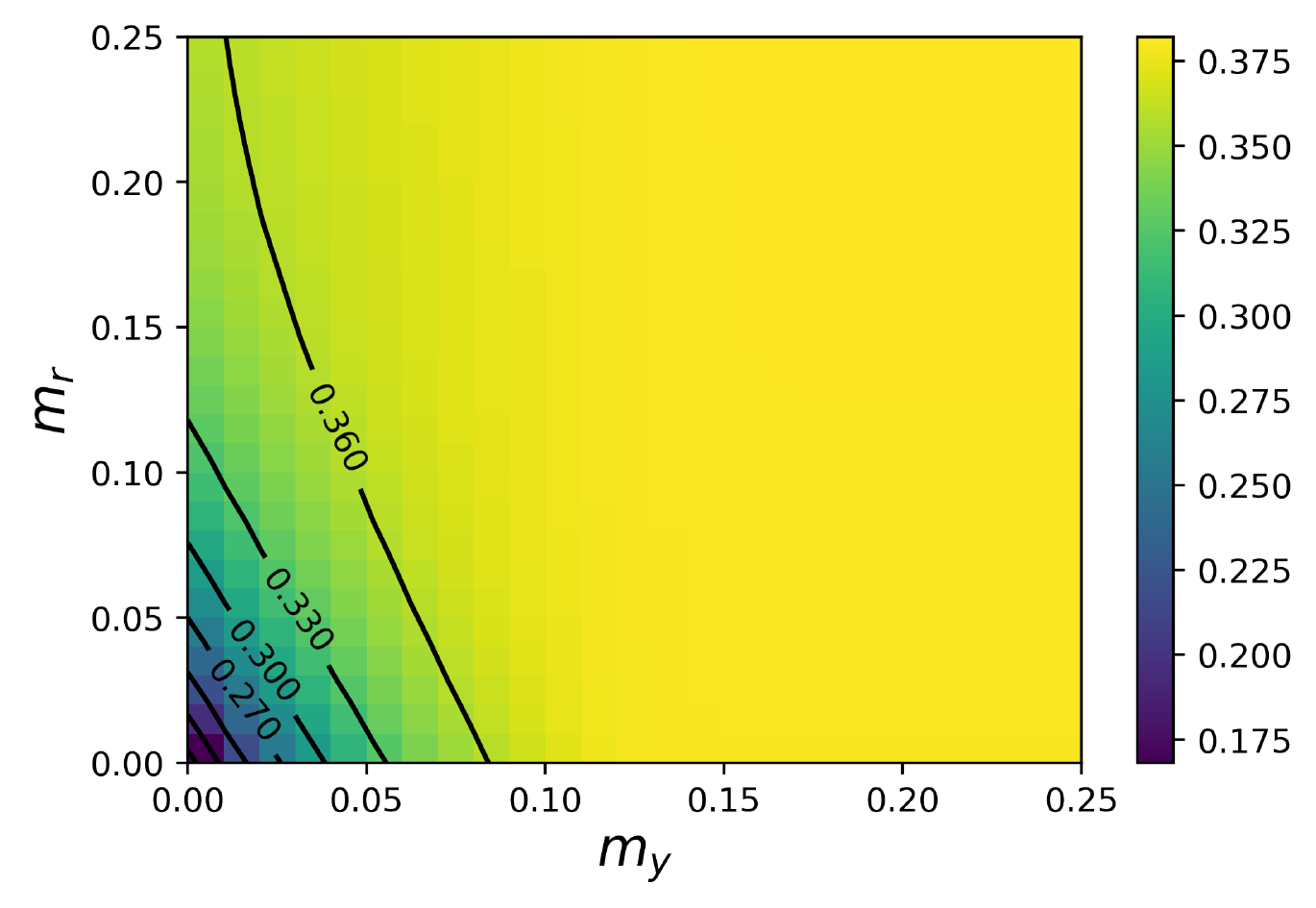}
        \caption*{(b) $r_{\max}=0.4$}
    \end{minipage}
    \begin{minipage}{0.24\linewidth}
        \includegraphics[width=\linewidth]{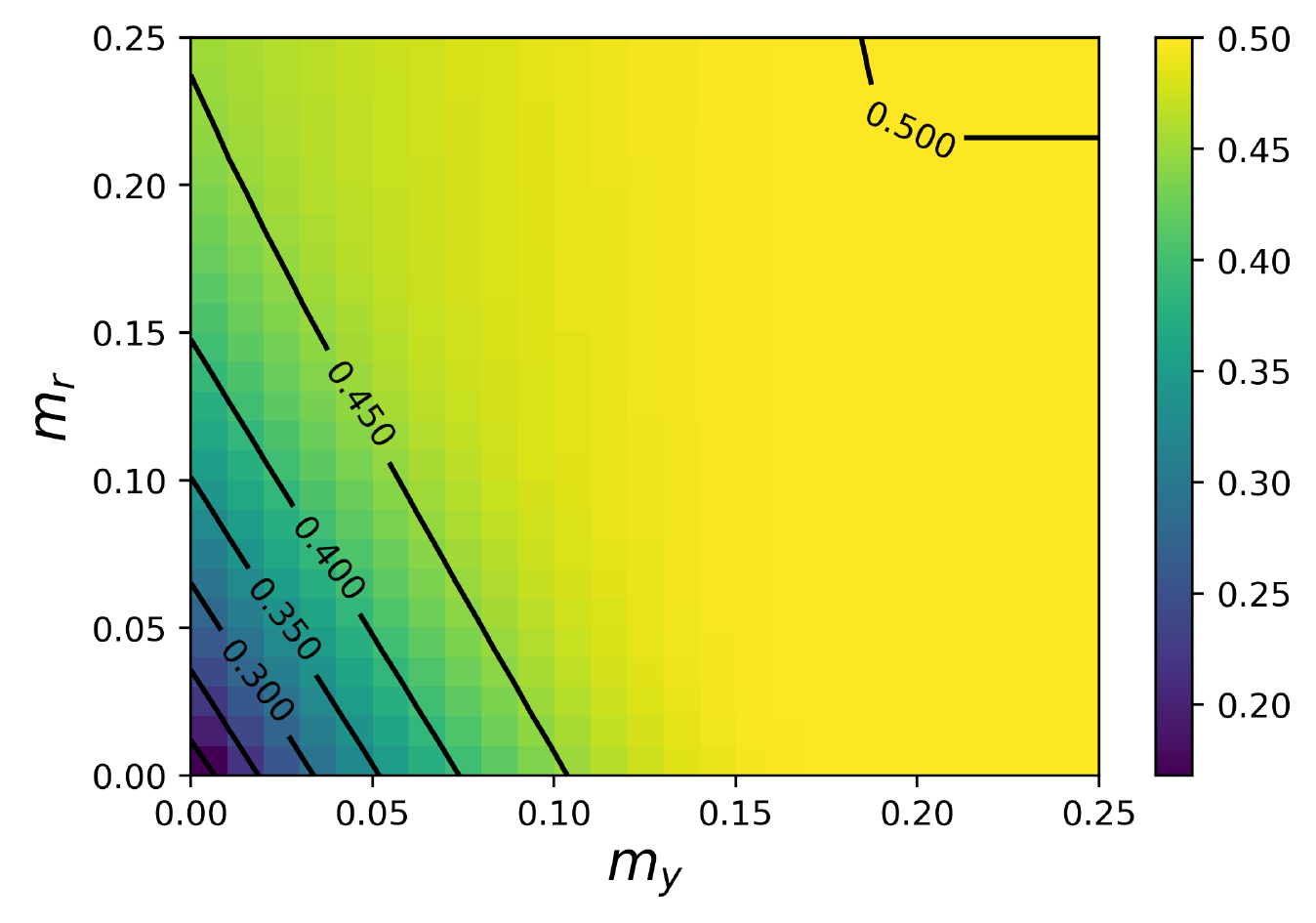}
        \caption*{(c) $r_{\max}=0.6$}
    \end{minipage}
    \begin{minipage}{0.24\linewidth}
        \includegraphics[width=\linewidth]{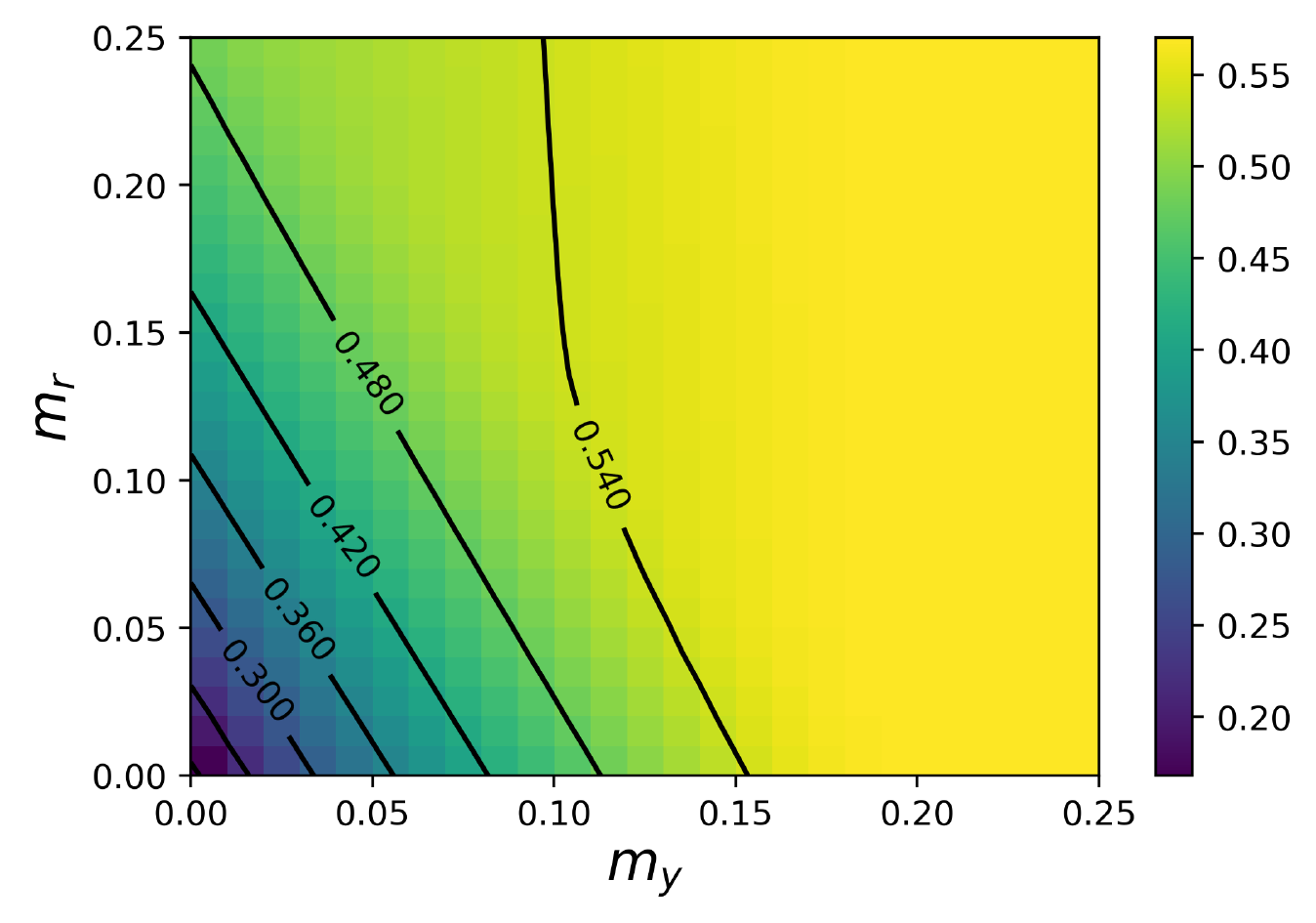}
        \caption*{(d) $r_{\max}=0.8$}
    \end{minipage}
    \caption{Heat map visualizations of $\Delta \bar y$ of synthetic data. Maximum fraction treated, $r_{\max}=$ (a) 0.2, (b) 0.4, (c) 0.6, and (d) 0.8. Lighter yellow colors correspond to larger change in the outcome,}
    \label{fig:syn_color}
\end{figure*}

\subsubsection{Affirmative Action-Constrained Interventions}
Alternatively, we can single out subgroups for preferential treatment, assigning different treatment ratios to subgroups within the leaf node $i$, which may improve the overall outcome for the entire population. We refer to this type of intervention as \textit{affirmative action} policy. For example, in the context of the school intervention program introduced earlier in the paper, affirmative action means that groups that benefit most from the treatment (have largest effect) should be preferentially assigned to the intervention. As another example, affirmative actions for  COVID-19 vaccinations means that minorities who are at high risk for COVID-19 complications should get priority access to early vaccines. 
To learn affirmative action interventions, we vary treatment ratios $r_{i, z=j}$ to maximize the overall outcome
\begin{equation}
    \bar y = \frac{1}{\sum_i n_i} \sum_i\sum_j n_{i, z=j}\cdot\left[
    r_{i,z=j}\cdot y_{i,z=j}^{(1)} + (1-r_{i,z=j})\cdot y_{i,z=j}^{(0)}
    \right]
\end{equation}
under the constraints:
\begin{itemize}
    \item We set an upper bound for the discrepancy of outcomes:
\begin{equation}
    \text{Bias}_y \leq m_y
\end{equation}
\item We set an upper bound for the amount of discrepancy in treatment opportunity we will tolerate as:
\begin{equation}
    \text{Bias}_r \leq m_r
\end{equation}
\item As before, we limit the number of individuals that can be treated
\begin{equation}
    \sum_i\sum_j n_{i,z=j}\cdot r_{i,z=j} \leq N_{\text{max}}^{(1)}
\end{equation}
\item And finally, all treatment ratios have to satisfy 
\begin{equation}
    0 \leq r_{i,z=j} \leq 1
\end{equation}
\end{itemize}

\subsubsection{Optimizing Fair Treatments}
Given the parameter of constrains, $(m_y, N_{\text{max}}^{(1)})$ or $(m_y, m_r, N_{\text{max}}^{(1)})$, we can use linear programming to solve for the optimal $\bar y$ and corresponding treatment assignment plan $r_i$ or $r_{i, z=j}$. 
The policies which are optimal under the constraints can be regarded as efficient policies.

\subsubsection{Boosting}
The causal tree learned from data depends on the random splitting of data into training, validation and estimation sets. Although this may not be a problem when we have sufficiently large dataset, random splits may cause instabilities when used for smaller datasets. To overcome this problem, we carry out multiple random spits of the data and train a causal tree for each data split. When designing an optimal policy, for every constraint parameter $(m_y, N^{(1)}_{\text{max}})$ or $(m_y, m_r, N^{(1)}_{\text{max}})$ we perform optimization for each of the causal tree trained and calculate the optimal outcome  as the average for all the causal trees. When boosting is involved, the treatment assignment can not be expressed using treatment probability in each leaf node, since we have multiple causal trees. 
Instead, we denote the optimal treatment assignment for causal tree with index $k$ as $r^{(k)}_{i,z=j}$, where $i$ is the index for the leaf node and $j$ is the index for values of $z$. 
Given the features $X$ and sensitive attribute $z$, in the case where affirmative action is allowed, we can define the treatment probability for the individual as
\begin{equation}
    r(X, z=j) = \frac{1}{N_{\text{tree}}}\sum_{k=1}^{N_{\text{tree}}} r^{(k)}_{l(X|k),z=j}.
    \label{eq:treat_individual}
\end{equation}
Here $N_{\text{tree}}$ is the number of causal trees trained and $l(X|k)$ is the leaf node index corresponds to an individual with feature $X$ in causal tree $k$. When affirmative action is not allowed, similarly we have
\begin{equation}
    r(X) = \frac{1}{N_{\text{tree}}}\sum_{k=1}^{N_{\text{tree}}} r^{(k)}_{l(X|k)}
    \label{eq:treat_individual_eq}
\end{equation}

\section{Results}

\subsection{Synthetic data}
As proof of concept, we demonstrate our approach on synthetic data representing observations from a hypothetical experiment.  The individual observations have features, $X = [x_0, x_1]$, $x_0, x_1 \sim U(0,1)$, drawn independently from a uniform distribution in range $[0,1]$. The treatment assignment and sensitive feature $z$ are generated independently using Bernoulli distributions: $z, t\sim Bernoulli(0.5)$. Finally, the observed outcomes $y$ depend on features and treatment as follows:
\begin{equation}
    y = t\cdot x_0 + 0.4\cdot t \cdot z \cdot x_0.
\end{equation}
Note that the feature $x_1$ is designed to not correlate with $y$.

Figure~\ref{fig:syn_data} shows the outcomes for the control ($t=0$) and treated ($t=1$) individuals. The two subgroups have the same outcome in the control case, but individuals from the protected subgroup ($z=1$) benefit more from the treatment ($t=1$), since their outcomes are higher than for individuals from the other group ($z=0$). 
Note that the larger the feature $x_0$, the larger the impact of treatment on the protected subgroup $z=1$. The disparate response to treatment creates a dilemma for decision makers---if both subgroups receive the same treatment ($\text{Bias}_r=0$), then higher population-wide outcome will be associated with a larger discrepancy in the outcomes for the two subgroups, hence, larger  bias ($\text{Bias}_y$). 

We train a causal tree to estimate the heterogeneous treatment effect using $X = [x_0, x_1]$. Given $6,000$ total observations, we use a third of the data for training the causal tree, a third for validation, and a third for estimation using honest trees \cite{athey-pnas16}. 
We estimate biases for the sensitive attribute $z$ and learn optimal interventions using data reserved in the estimation set. 

\begin{figure}[th]
    \centering
    \begin{minipage}{1.0\linewidth}
        \includegraphics[width=\linewidth]{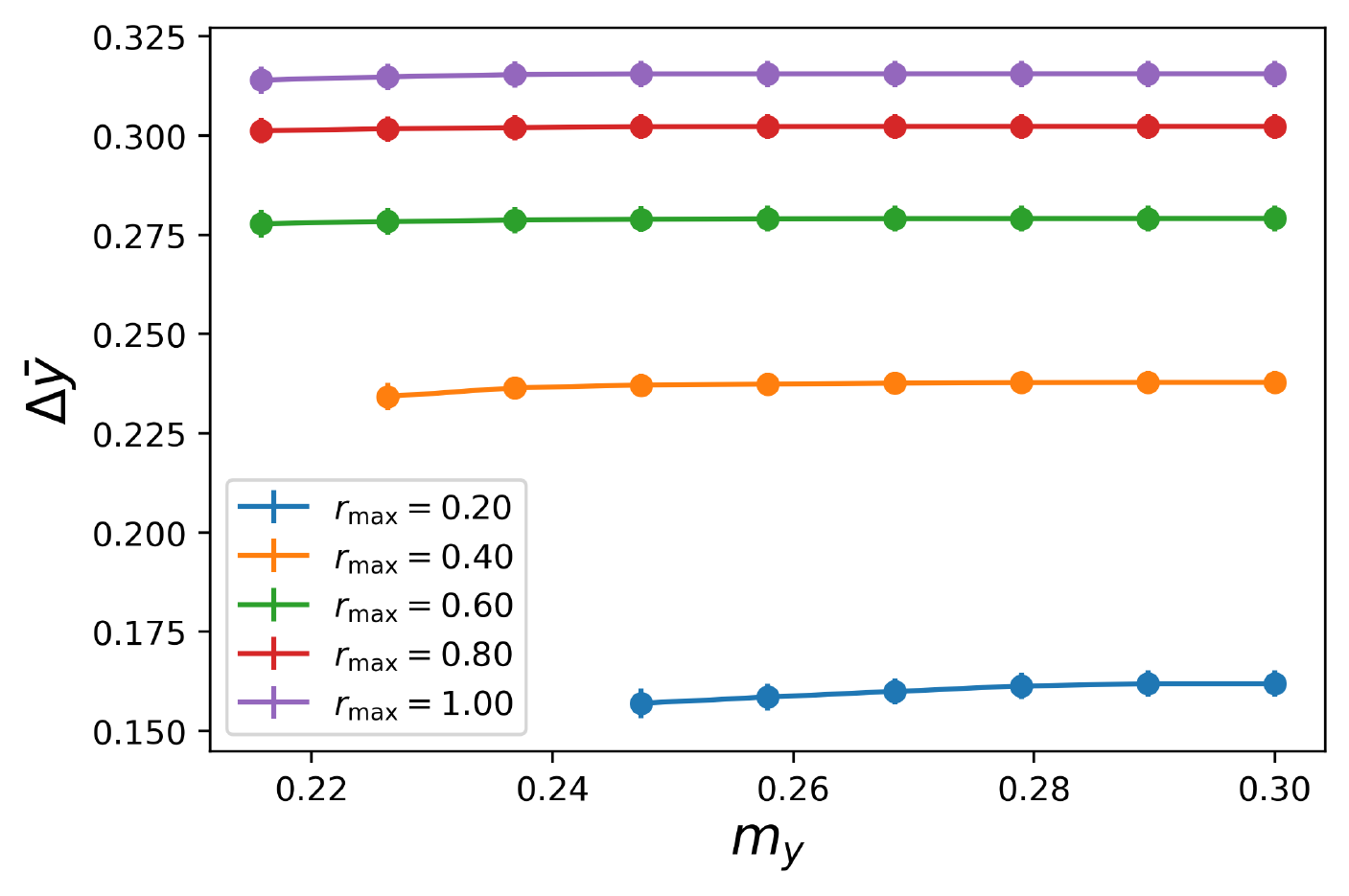}
        \caption*{(a) Equal opportunity }
    \end{minipage}
    \begin{minipage}{1.0\linewidth}
        \includegraphics[width=\linewidth]{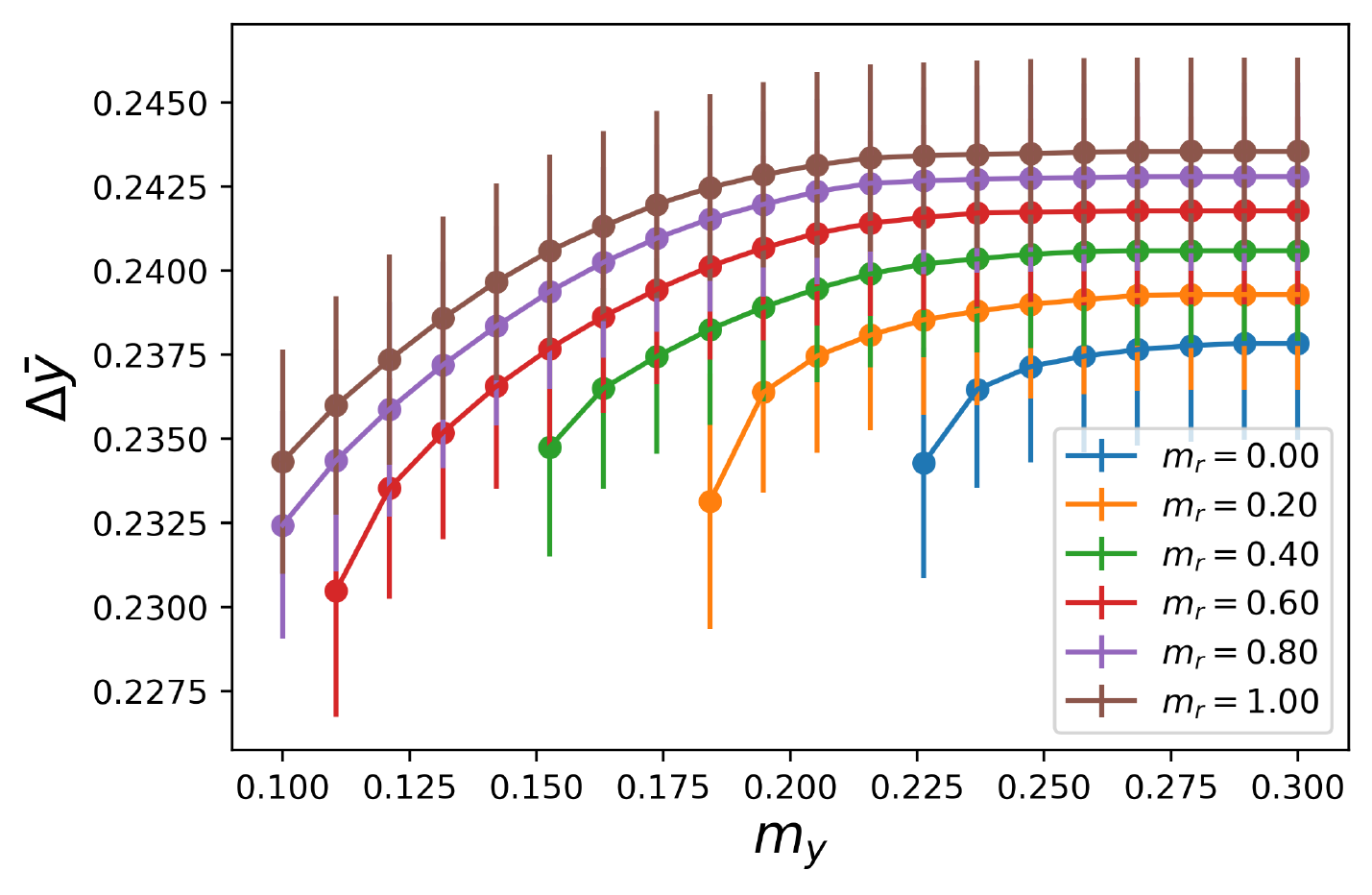}
        \caption*{(b) Affirmative action}
    \end{minipage}
    \caption{Performance improvement of different treatment policies in EdGap data. (a) Performance improvement due to an intervention $\Delta \bar y$ vs. $m_y$ (allowed mean score difference between counties with high/low Black household ratio) when equal treatment opportunity is assumed. Different curves shows different $r_{\max}$, the maximum treatment ratio. The leftmost point of each curve shows the edge of the infeasible region. (b) $\Delta \bar y$ vs. $m_y$ when affirmative action is allowed. Here $r_{\max}=0.4$ and different curves show results with different constrains on affirmative action $m_r$.}
    \label{fig:edgap_bound}
\end{figure}

\noindent \paragraph{Equal Treatment Policy}
First we consider the equal treatment policy, where individuals from either subgroup are equally likely to be treated. As described in the preceding section, in this case $\text{Bias}_r = 0$. To model limited resources, such as limited doses of a vaccine or limited number of  spots in the academic intervention program, we assume that we can only treat up to $N^{(1)}_{\text{max}}$ individuals.
For simplicity, we introduce 
$r_{\text{max}} = {N^{(1)}_{\text{max}}}/{N}$, which is the maximum treatment ratio as a measure of resource limit.
Also we use $$\Delta \bar y = \bar y - \frac{\sum n_i\cdot y_i^{(0)}}{\sum n_i},$$ as a measure of the \textit{improvement of the outcome} after treatment.
We vary treatment ratio $r_{\text{max}}$ in $\{0.2,0.4,0.6,0.8,1.0\}$, and for each value of $r_{\text{max}}$ plot the improvement in overall outcome of the intervention ($\Delta \bar y$) as a function of $m_y$, the upper limit of the bias in outcomes ($\text{Bias}_y$). Figure~\ref{fig:syn_bound}(a) shows that as we treat more individuals (larger $r_{\text{max}}$), there is greater benefit from the intervention in terms of larger overall outcome ($\Delta \bar y$). Additionally, as we tolerate more bias ($m_y$ increases), the overall outcome also increases. However, for large enough  $m_y$, there is no more benefit from the intervention. 
In this case, we have assigned all the necessary treatment and allowing more bias will not further improve the outcome. In other words, when no more people can be treated under the constraint $r_{\text{max}}$, and $\text{Bias}_y$ is maximized. 

\noindent \paragraph{Affirmative Action Policy}
To see how affirmative action could improve the average overall outcome, we fix $r_{\text{max}} = 0.8$ and vary $m_r$ in $\{0, 0.05, 0.10, 0.15, 0.20, 0.25\}$. This allows us to prioritize protected subgroups for treatment. As more individuals from protected subgroups are treated, the treatment ratios become different, increasing $\text{Bias}_r$. 

Figure~\ref{fig:syn_bound}(b) shows the overall outcome $\bar y$ as a function of maximum bias $m_y$ for different values of $m_r$.  The curve $m_r = 0$ is the degenerate case where the previous policy of equal opportunity holds. We see that at large $m_y$, different curves of different values of $m_r$ reach the same upper bound of $\bar y$, which is constrained by $r_{\text{max}}=0.8$. For lower values of $m_y$, affirmative action dramatically increases $\bar y$, i.e., preferentially selecting individuals from protected subgroups for treatment increases the overall benefit of treatment.

\begin{figure*}[tbph]
    \centering
    \begin{minipage}{0.24\linewidth}
        \includegraphics[width=\linewidth]{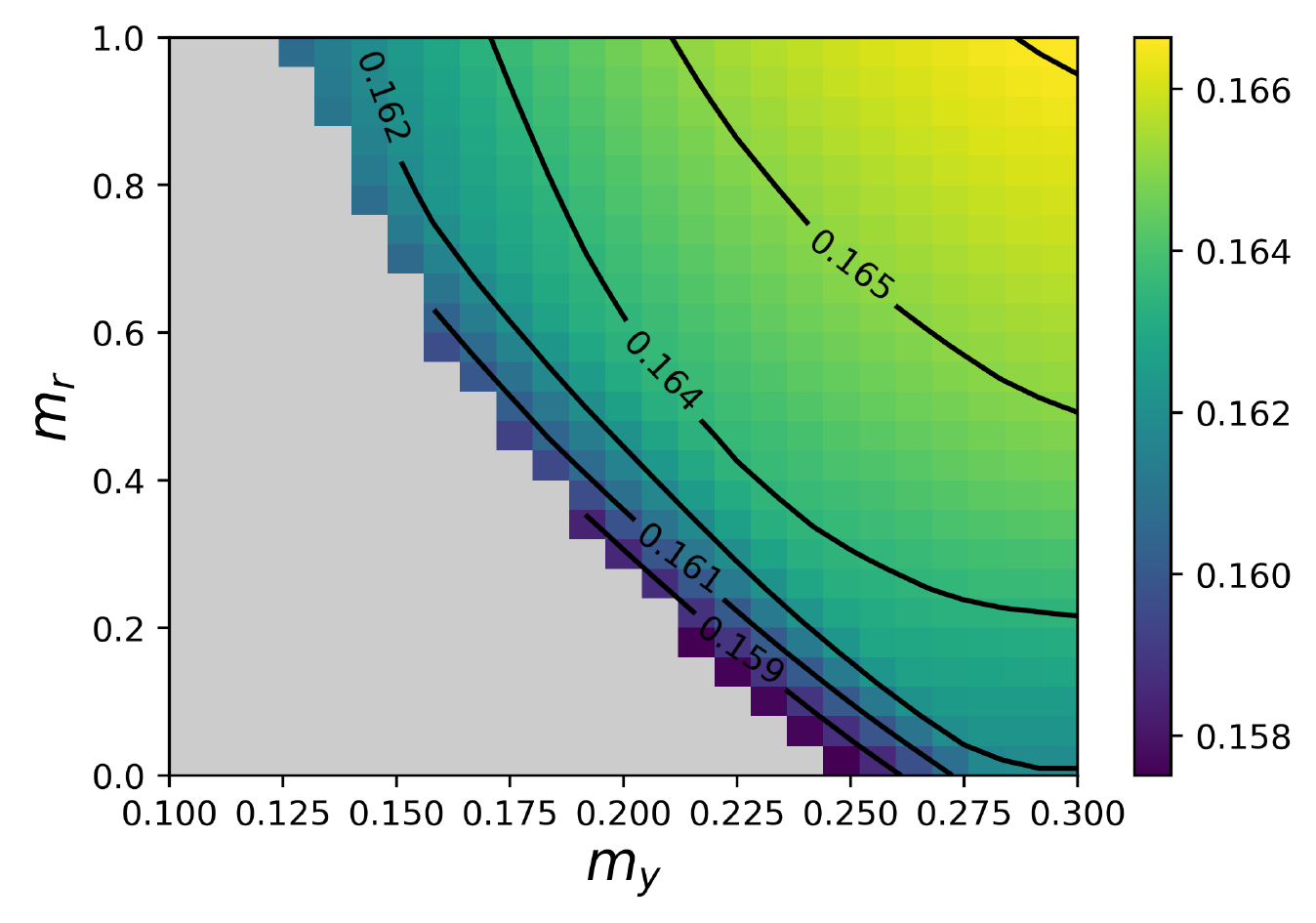}
        \caption*{(a)}
    \end{minipage}
    \begin{minipage}{0.24\linewidth}
        \includegraphics[width=\linewidth]{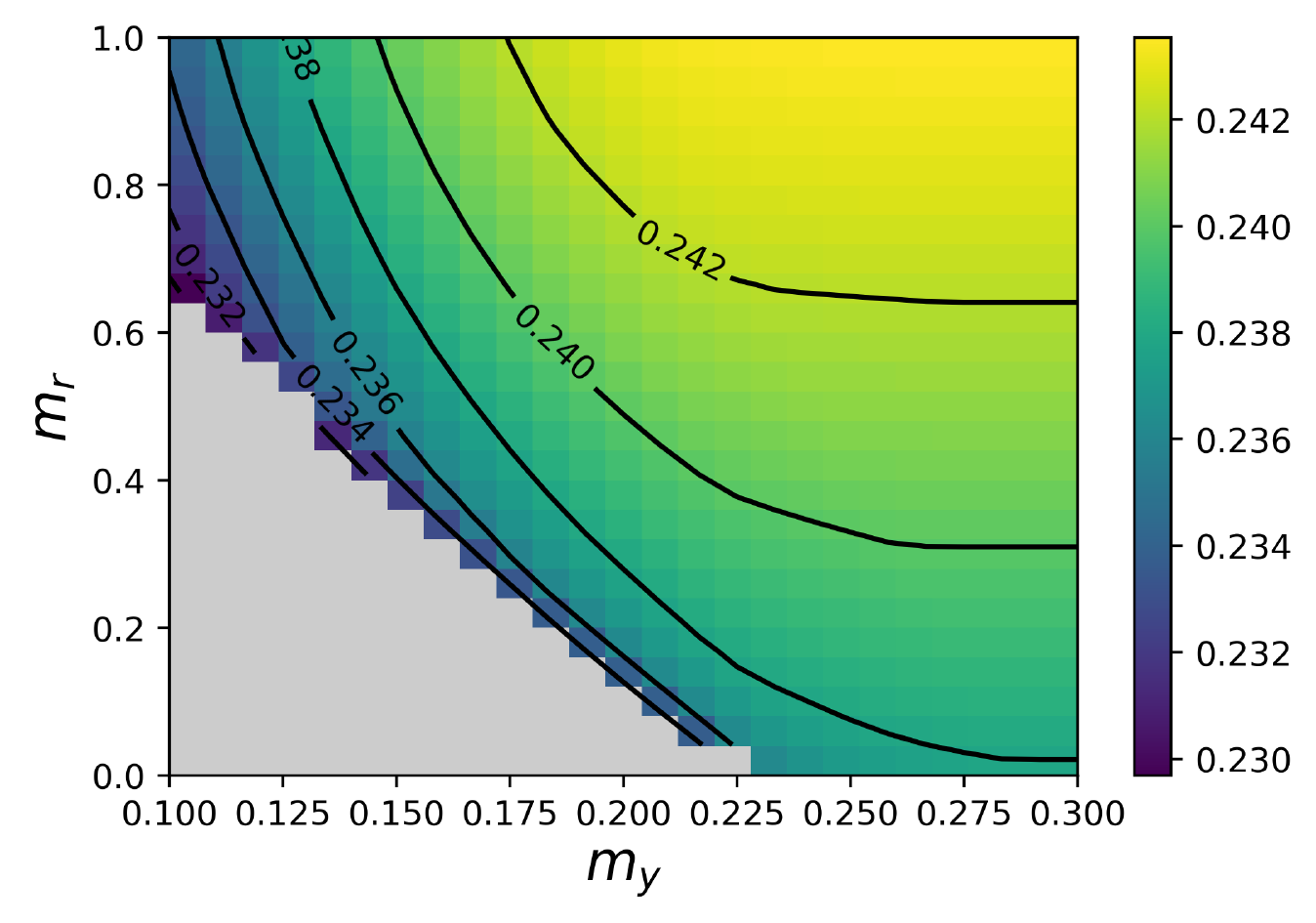}
        \caption*{(b)}
    \end{minipage}
    \begin{minipage}{0.24\linewidth}
        \includegraphics[width=\linewidth]{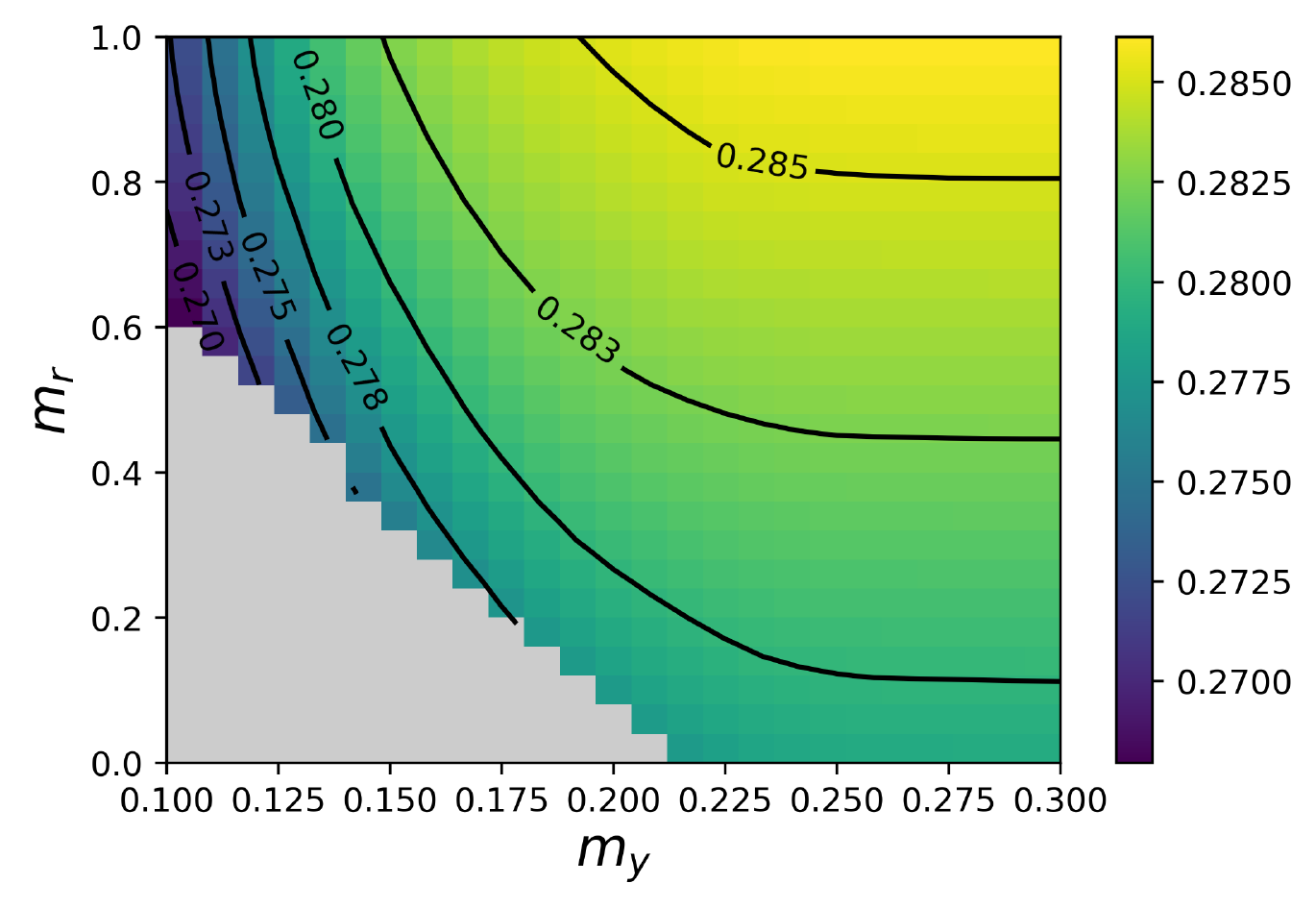}
        \caption*{(c)}
    \end{minipage}
    \begin{minipage}{0.24\linewidth}
        \includegraphics[width=\linewidth]{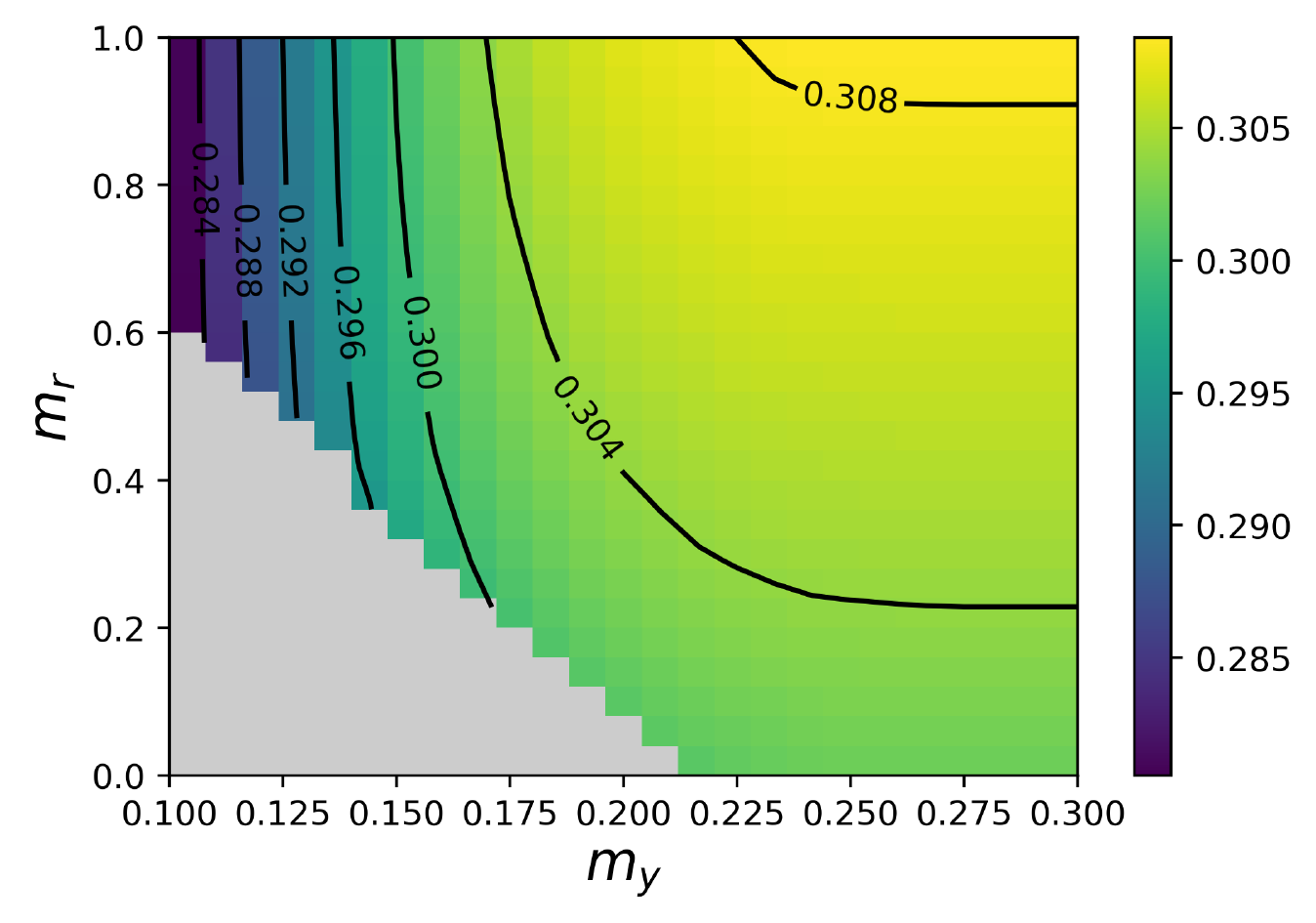}
        \caption*{(d)}
    \end{minipage}
    \caption{Heat map visualizations of performance improvement $\Delta \bar y$. Maximum fraction treated, $r_{\max}=$ (a) 0.2, (b) 0.4, (c) 0.6, and 0.8. Lighter colors correspond to greater overall benefits of treatment, while the infeasible region is shown as grey.}
    \label{fig:edgap_heat}
\end{figure*}

\begin{figure*}[tbph]
    \centering
    \begin{minipage}{0.49\linewidth}
        \includegraphics[width=\linewidth]{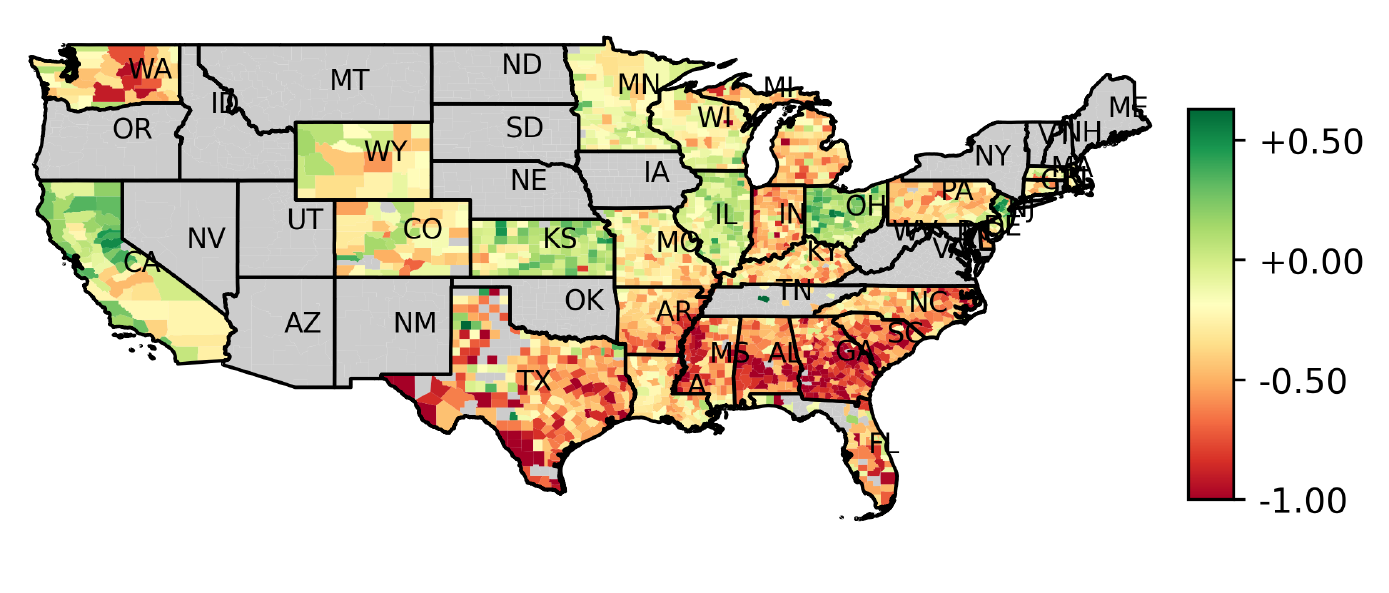}
        \caption*{(a) Test scores}
    \end{minipage}
    \begin{minipage}{0.49\linewidth}
        \includegraphics[width=\linewidth]{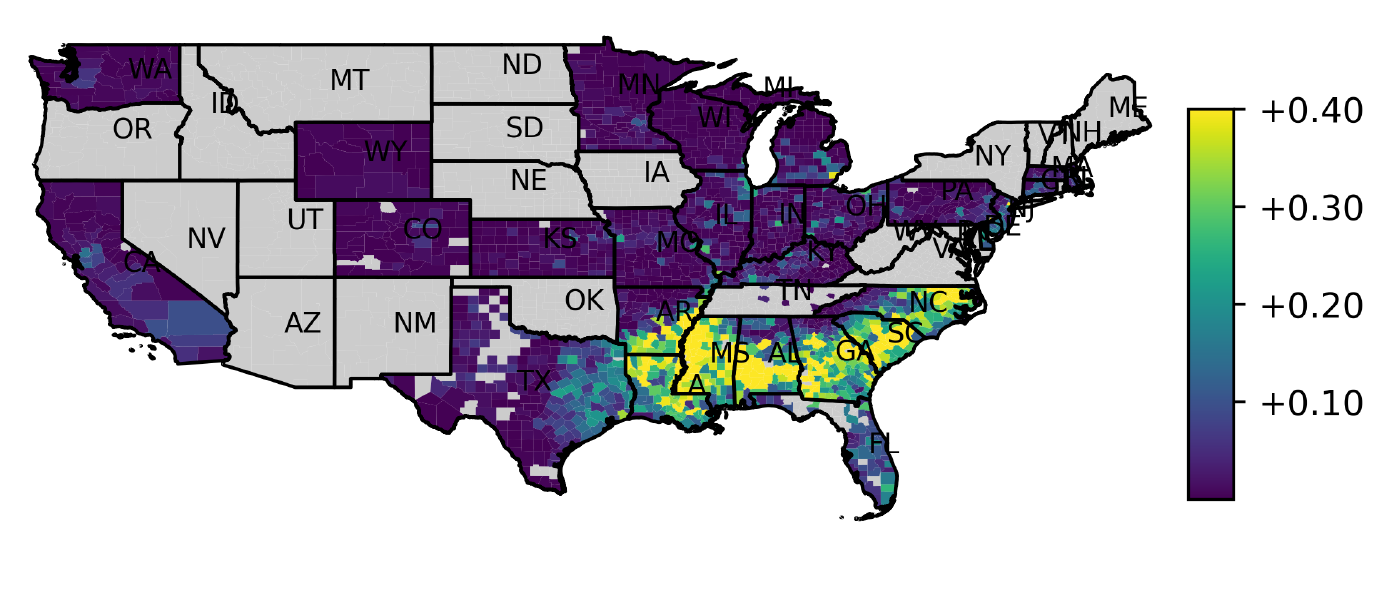}
        \caption*{(b) Population African-Americans}
    \end{minipage}    \begin{minipage}{0.49\linewidth}
        \includegraphics[width=\linewidth]{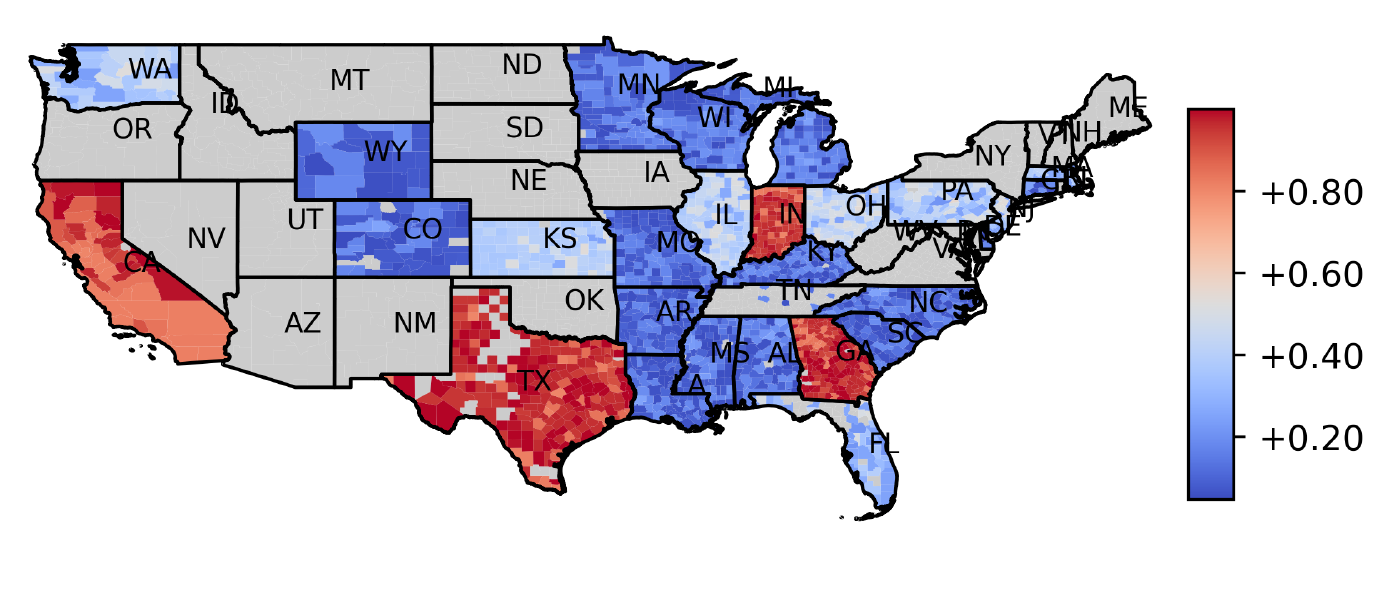}
        \caption*{(c) Equal opportunity policy}
    \end{minipage}
    \begin{minipage}{0.49\linewidth}
        \includegraphics[width=\linewidth]{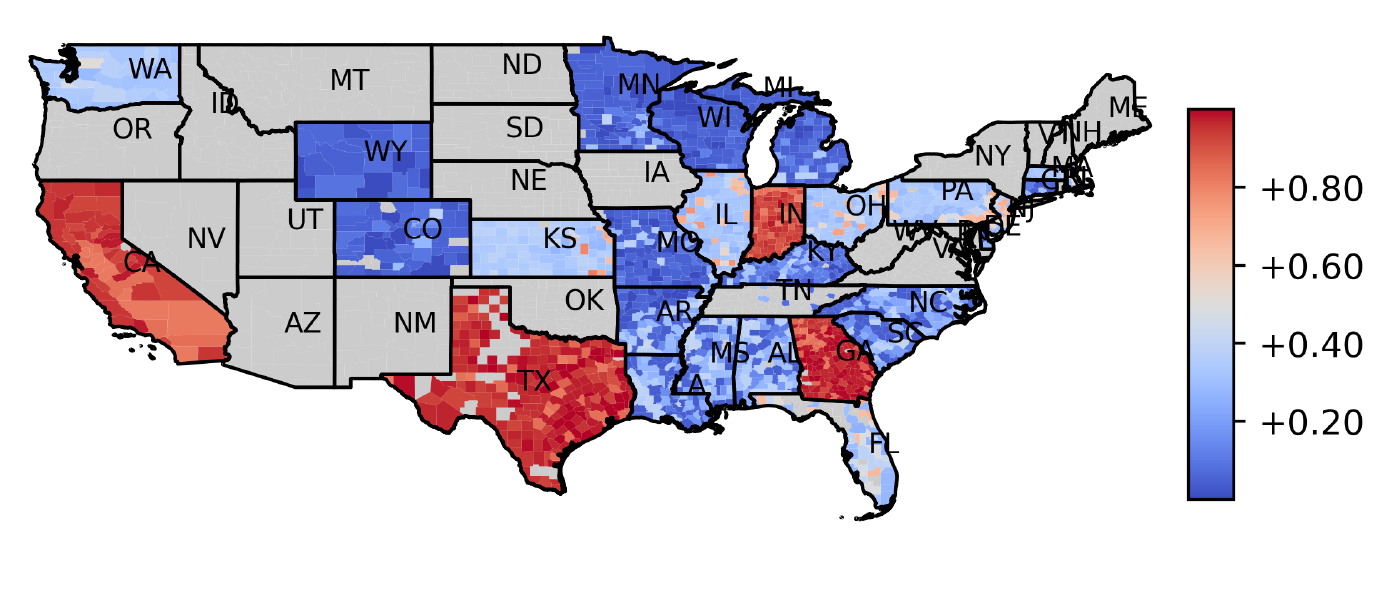}
        \caption*{(d) Affirmative action policy}
    \end{minipage}
    \caption{EdGap test scores data. (a) Geographical distribution of mean test score. (For visualization, the z-score is plotted, value clipped at -1.0 and 1.0.) (b) Geographical distribution of Black household ratio. (For visualization, the values are clipped at 0.4 or 40\%.) (c) Learned optimal treatment assignment when equal treatment opportunity is assumed ($m_y = 0.25, r_{\max}=0.40, m_r = 0$). (d) Learned optimal treatment assignment when affirmative action is allowed ($m_y = 0.25, r_{\mathrm{max}}=0.40, m_r = 1.0$).}
    \label{fig:edgap_map}
\end{figure*}

\noindent \paragraph{Trade-offs between $m_r$ and $m_y$}
To further illustrate the how $m_y$ and $m_r$ affect the outcome $\bar y$, we plot $\bar y$ against $m_y$ and $m_r$ using heat map and contour line, as shown in Fig. \ref{fig:syn_color}. 
The heatmap and contour lines demonstrates the trade-offs between the two biases. In order to maintain the same level of benefit from the intervention (moving along the contour lines) while reducing maximum allowed treatment opportunity bias $m_r$ requires us to tolerate larger treatment outcome bias $m_y$ and vice versa.

\subsection{EdGap Data}
The \textit{EdGap} data contains education performance of different counties of United States. The data we used contains around 2000 counties and 19 features. The features include funding, normalized mean test score, average school size, number of magnet schools, number of charter schools, percent of students took standardized tests and average number of students in schools receiving discounted or free lunches. Besides these features, we  have census features for each county including household mean income, household marriage ratio, Black household ratio, percent of people who finished high school, percent of people with a bachelor's degree, employment ratio, and Gini coefficient. We use z-score normalized mean test score as the outcome. We  binarize school funding and the county ratio of Black households to be above and below the median values as treatment indicator and sensitive feature, respectively. In summary, we are interested in the heterogeneous effect of funding increase on different counties and we want to design a fair intervention which reduces the education performance difference between Black and non-Black populations.

We use one third of data as training, validation and testing, respectively and train forty boosted causal trees and report the average performance. We first show the result where equal treatment opportunity is assumed. We plot the overall mean score after treatment 
versus $m_y$ in Fig.\ref{fig:edgap_bound}(a).  
Unlike the synthetic data (Fig.\ref{fig:syn_bound}(a)), we find an infeasible region (note that the left bound of curves are different; beyond the left bound is the infeasible region). The lower the maximum allowed treatment ratio, the larger the infeasible region. 
This is because, without any treatment, there is a difference in the mean of average test scores for county with more Black population and less Black population. If the constrain $m_y$, score difference between two groups of county, is set to be too low and the maximum allowed treatment ratio $r_{\text{max}}$ is also low, the constraints cannot be satisfied. But on the other hand, we also notice that if affirmative action is allowed, we can assign more counties with high Black ratio to treatment and dramatically improve the mean outcome and also reduce the infeasible region (allow greater fairness), as shown in Fig. \ref{fig:edgap_bound}(b). We also plot $\Delta \bar y$ versus $m_y$ and $m_r$ for different $r_{\text{max}}$ in heat maps (Fig. \ref{fig:edgap_heat}). We observe that the size of infeasible region (grey region) is reduced as $m_r$ and $r_{\text{max}}$ increase.

To further understand the bias in the data and the fair intervention we learned, we visualize the geographical distribution of data and the learned treatment assignment in Fig. \ref{fig:edgap_map}. 
We first plot the mean test score of counties and the ratio of Black household in Fig. \ref{fig:edgap_map}(a)--(b), respectively. We see that in the southeast states, from Louisiana to North Carolina, there are counties with high ratio of Black households. Correspondingly, we also see that the mean test scores of those counties are lower than national average due to chronic under-funding and racism. To illustrate the effect of affirmative action, we plot the learned optimal treatment assignments of two sets of parameters. First, we consider the case where we assume equal treatment opportunity. We use parameters $m_y = 0.25$, $r_{\text{max}}=0.4, m_r = 0$. Then for the case where affirmative action is allowed, we use parameters $m_y = 0.25$, $r_{\text{max}}=0.4, m_r = 1.0$. For both plots, we can see that counties in California, Texas, and Georgia have a high probability of being assigned to treatment. This is because the causal tree model predicts that counties in those state have higher treatment effect $\tau(X)$. Importantly, comparing Fig.\ref{fig:edgap_map}(c) and (d), we see when affirmative action is allowed, the counties in Louisiana, Mississippi, Alabama, South Carolina, and North Carolina have a high probability of being assigned to treatment. 
The treatment will not only improve the overall performance, but will also reduce performance difference between counties with high and low Black households.

\section{Discussion}
In this paper, we ask how we can learn intervention policies that both improve desired outcomes and increase equality in treatment across any number of protected classes. To do so, we first create novel metrics to quantify the fairness of any policy, and then create fairer policies based on two complimentary, but distinct, definitions of fairness. These findings demonstrate a trade-off between policies that maximize outcomes and fairness. Increasing the overall outcome can bring unintended unequal treatments between protected classes. That said, the ways to mitigate this unfair treatment has its own trade-off. Policies that provide equal treatment to all classes still provide substantial overall unequal treatment. Affirmative action policies, in contrast, provide greater overall fairness, but imply that subgroups must receive unequal treatment. Finally, we provide an algorithm that offers the best policies, conditional on the trade-offs policy-makers' desire.

While this methodology offers substantial benefits to policy-makers, our work still has limitations. First, the algorithm and metrics are tailored to causal trees. While trees are highly interpretable, numerous other causal methods exist, and other algorithms need to be tailored to these other methods \cite{wager-jasa17,athey-annals19,Kunze2019}. Second, there is an open question of how Bayesian networks \cite{pearl-book09}, which model the pathways of causality, relate to algorithms that model heterogenous treatment effects. Future work must explore how fair policies created via causal models relate to potentially fair policies created by Bayesian networks.

\section{Acknowledgements}
This project has been funded, in part, by DARPA under contract HR00111990114.

\bibliographystyle{plain}
\bibliography{ref}

\end{document}